\pdfoutput=1
\documentclass[10pt,twocolumn,letterpaper]{article}
\usepackage[T1]{fontenc}
\usepackage[pagenumbers]{cvpr}
\usepackage{amsmath,amssymb,booktabs,graphicx,xcolor}
\usepackage[pagebackref,breaklinks,colorlinks]{hyperref}
\def\confName{CVPR}\def\confYear{2027}
\title{Seeing Is Not Enough: Knowledge-Grounded Active Inquiry from Uncertain Visual Evidence}
\author{Yucheng Liu\\
Department of Engineering Science, University of Oxford\\
{\tt\small yucheng.liu25@gmail.com}
\and
Everett X. Wang\thanks{Corresponding author.}\\
Beijing Institute of Nanoenergy and Nanosystems, Chinese Academy of Sciences\\
{\tt\small wangxiaolin@binn.cas.cn}}
\begin{document}\maketitle
\begin{abstract}
Similar tongue findings can accompany different syndromes in traditional Chinese medicine (TCM), with decisive symptoms absent from the photograph. We present a framework for tongue assessment in which image and pulse findings guide targeted symptom inquiry and acquired answers refine the syndrome shortlist before reporting.

Image/pulse predictions guide selection from a sourced symptom question bank. Known answers update the assessment used to select the next question. A calibrated loss of support for the initial prediction triggers reassessment; a subsequent report requires fresh answers supporting a stable candidate. Question and answer records make these decisions inspectable.

We construct 2,600 cases from distinct real tongue photographs, assigned syndromes, simulated symptom responses and rendered pulses; 2,599 enter five-fold source-disjoint evaluation and one is training-only. On clean images, image/pulse prediction alone achieves 46.4\% fine-syndrome accuracy. At acquisition cost 12, inquiry without image/pulse evidence reaches 34.4\%, whereas their combination reaches 65.8\% accuracy and 90.8\% top-three inclusion. A same-prior, training-optimized static order reaches 59.4\% and 87.3\%, respectively. In a 200-case visual-training intervention within the same bank, the first question changes in 46.7\% of cases and the inquiry NLL--cost curve improves. Component comparisons distinguish these acquisition gains from the coverage--cost tradeoff of reassessment. These controlled results support image-guided acquisition of missing symptoms for ranked TCM decision support.
\end{abstract}

\section{Introduction}
Tongue assessment in traditional Chinese medicine (TCM) combines visible findings with symptoms that a photograph cannot show. Similar coating, colour, shape, or cracks can support several possible syndromes, leaving the assessment dependent on what is asked next. Computer vision can help identify which missing symptom would distinguish those possibilities. The answer must then be able to change the assessment, especially when it weakens the initial interpretation.

Existing TCM systems address important parts of this process. TongueVLM generates tongue descriptions, reasons about constitution and supports multiround dialogue~\cite{peng2026}. LingLan integrates tongue, pulse and inquiry descriptions into structured records for diagnostic prediction~\cite{chen2026}. Li et al. select inquiry questions from candidate syndromes~\cite{li2022}. Their evaluations leave open how much learned image/pulse evidence improves the acquisition of missing symptoms at a fixed cost, and how conflicting answers should change report/review decisions. We focus on these questions when visual findings are ambiguous or image degradation favours an incorrect initial candidate.

\begin{figure*}[t]
\centering\includegraphics[width=\textwidth]{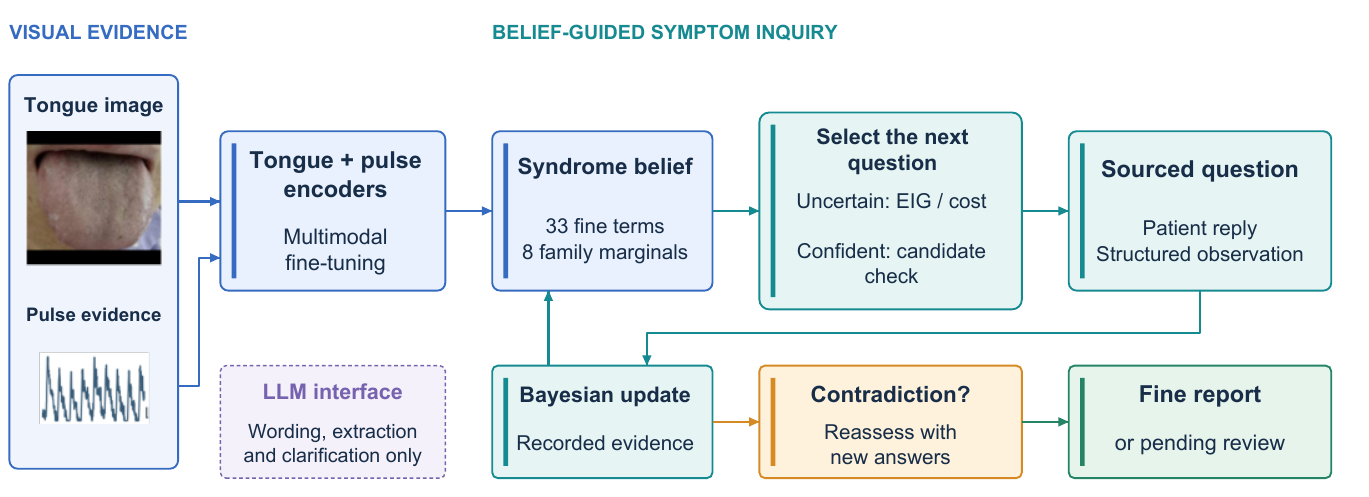}
\caption{Visual evidence enters a feedback process. Separately pretrained tongue and pulse encoders are fine-tuned with the multimodal classifier. Their calibrated fine posterior controls sourced symptom acquisition. A candidate check tests a confident candidate; a calibrated drop in support for the initial image/pulse prediction starts reassessment. The Bayesian state changes only from recorded structured observations. The language interface supplies phrasing, extraction, and clarification. A pending discrepancy closes only after fresh supporting evidence meets the declared criterion.}
\label{fig:pipeline}
\end{figure*}

We connect tongue-image and pulse findings to a sourced symptom question bank over 33 syndromes within eight families (Figure~\ref{fig:pipeline}). The current assessment guides the next question, and each known structured answer revises the syndrome shortlist. Questions can test a likely candidate against alternatives that initially received little support. When support for the initial image/pulse prediction drops past a calibrated threshold, the system continues inquiry and requires fresh supporting answers before reporting. Cases with unresolved discrepancies remain under review. Each question, answer and change in the assessment is recorded.

We contribute a framework and controlled evaluation for visually guided symptom inquiry in TCM. Matched-cost comparisons measure how image/pulse evidence and adaptive questioning improve the syndrome shortlist. A visual-training intervention isolates how image representation changes subsequent questions, while prior interventions distinguish family from within-family information. Reporting experiments measure precision, coverage and cost when a discrepancy requires fresh supporting answers before report release.

The benchmark contains 2,600 constructed cases with real tongue photographs, assigned syndromes, simulated symptom responses and rendered pulses; 2,599 are evaluated across five source-disjoint folds and one is training-only. On clean images, combining image/pulse evidence with inquiry reaches 65.8\% fine accuracy and 90.8\% top-three inclusion at cost 12, compared with 46.4\% fine accuracy before inquiry and 34.4\% for inquiry without image/pulse evidence. A static order with the same prior reaches 59.4\%. Shared cases and answer realizations make these paired tests of the value of evidence and its acquisition.

\section{Related work}
Active feature acquisition balances prediction quality against the cost of obtaining observations. Li and Oliva learn generative surrogate models to guide acquisition~\cite{li2021}; Valancius et al. use an acquisition-conditioned oracle to account for jointly informative features~\cite{valancius2024}. White et al. select visually grounded clarification questions by expected information gain~\cite{white2021}. In pathology, PathFound revisits diagnostic hypotheses through targeted evidence seeking~\cite{hua2026}. We examine how the visual representation affects nonvisual question selection and how answers that contradict the initial prediction should affect reporting.

Related TCM systems connect assessment with consultation. TongueVLM evaluates descriptions, constitution reasoning and multiround dialogue through response agreement~\cite{peng2026}; these endpoints do not measure symptom acquisition at matched cost. LingLan evaluates prediction from structured records combining tongue, pulse and inquiry descriptions~\cite{chen2026}, with the inquiry information already supplied. Li et al. evaluate candidate-driven inquiry in simulated consultations, using sampled candidate sets to represent preceding examinations~\cite{li2022}. The household robot proposed in patent application CN120544858A also uses tongue findings to adjust Bayesian questioning~\cite{patent2025}. Our study tests the acquisition value of learned image/pulse beliefs and measures report precision, coverage and cost after conflicting answers. These are controlled internal comparisons; the cited systems are not rerun as baselines.

Calibration and selective prediction provide tools for deciding when to report. Temperature scaling adjusts predictive confidence~\cite{guo2017}, and SelectiveNet learns prediction and rejection jointly~\cite{geifman2019}. Our thresholds are calibrated on separate cases and applied as answers accumulate. We measure report precision and coverage together. After a contradiction, a candidate must also gain support from new answers.

\section{Benchmark and data separation}
\subsection{Real photographs and assigned case state}
The photograph source is the CC0 Dryad release \emph{TMC-Tongue}~\cite{dryad2026}. Its local inventory contains 6,719 photographs and annotation IDs 0--19. Twelve supported visual labels supervise the tongue encoder; healthy and anatomical-zone labels are excluded. Full-image, crop, and mirrored-crop duplicate comparisons form conservative transitive source clusters. Each allocated photograph represents a distinct source cluster.

The structured knowledge base defines eight families and 33 fine terms. We match each case to a distinct photograph whose annotations support the requested findings, exclude conflicting matches, and record the assignment. Source annotations, requested findings, and encoder predictions are separate records. Visual inputs are an unmasked RGB crop and seven colour measurements computed using an image-derived mask; source annotations and assigned labels are not input features. Paleness is an image-derived colour-rank proxy fitted using the separate tongue training pool. Unannotated properties remain unknown.

The benchmark contains 2,600 constructed cases matched from the photograph source bank. Each case contains its assigned fine term and family, photograph and mask, source provenance, pulse array and visualizations, private symptom state, fixed answer draws, split assignment, and hashes. All 2,600 case photographs are disjoint from the 1,200 photographs used for tongue pretraining and development. A common case identifier links these records in the casebook.

FT019 has only one photograph satisfying its supported required findings. It is used only for training; augmented views share its source identifier and likelihood record. The classifier retains all 33 candidates. Thus the benchmark contains 2,600 constructed cases: 2,599 evaluated cases across 32 represented fine terms and one training-only case. Evaluation folds contain 520, 520, 520, 520 and 519 cases; term counts reflect matching capacity.

\subsection{Pulse and symptom observations}
Pulse evidence is rendered in the format of a multi-site, multi-pressure sensor array. Its numerical tensor has shape $2\times3\times3\times512$: two hands, three wrist sites, three pressure levels, and 512 time samples at 64\,Hz. The classifier consumes the resulting 18 channels. A grey-scale heatmap shows amplitude over channel and time, with gaps separating channel groups. Waveform plots display the same numerical signal. Pulse-classification work motivates using waveform and frequency features~\cite{huang2020}. The generator represents 28 pulse qualities.

Each case has a fixed symptom state. The answer endpoint returns the selected concept as absent, mild, marked, or unknown. Unknown answers incur cost and leave the posterior unchanged. All methods use the same ten answer draws for each case. Seven separate structured inputs support safety screening and are excluded from the syndrome likelihood and diagnostic question count.

\begin{table}[t]
\centering\caption{Data allocation. Counts denote distinct photographs or independently generated pulse examples.}
\label{tab:data}
\begin{tabular}{lrrr}\toprule
Stage & Train & Cal./dev. & Eval.\\\midrule
Tongue pretraining &1,000&200&--\\
Pulse pretraining &800&200&--\\
Visual-stage study &600&200&200\\
Each cross-fit fold &${\sim}1,560$&${\sim}520$&${\sim}520$\\\bottomrule
\end{tabular}
\end{table}

\subsection{Training, calibration, and evaluation}
Five outer folds rotate three folds for model and answer-likelihood training, one for calibration, and one for evaluation. Each nonsingleton case is evaluated exactly once. Each analysis fixes its settings across all five folds before that analysis is evaluated. Evaluated source clusters are excluded from model fitting and calibration within each fold. The visual-stage intervention uses a 600/200/200 train/development/evaluation allocation within the same bank (Table~\ref{tab:data}); its evaluation cases also occur in the cross-fit study, so it is not an independent replication.

Within each experiment, compared methods share case identities, likelihood-training partitions, answer draws, and acquisition costs. Supporting records provide protocols, input hashes and comparison outcomes, including configurations that do not meet the reporting criterion. Models and question sequences accompany the component, prior, generator and language analyses. These analyses were developed retrospectively on the same source bank.

\section{Image-guided symptom inquiry and reporting}
\subsection{Objective multimodal prediction}
An ImageNet ResNet-18~\cite{he2015} is first fine-tuned on the 12 visual labels using 1,000 tongue photographs. A separate pulse encoder learns 28 operational pulse qualities from 800 training and 200 development examples. The multimodal representation concatenates 512 tongue features, 128 pulse features, and seven image-derived colour measurements. A fine classifier produces $v(h\mid x,z)$ for $h\in\{1,\ldots,33\}$, where $x$ and $z$ are the tongue and pulse observations. Family probabilities sum fine probabilities within the fixed hierarchy.

The neural head combines projected tongue, pulse, and colour features with a 33-way output. Training uses fine-label cross-entropy with label smoothing and an auxiliary family loss. The full fine-tuning arm updates every tongue and pulse encoder stage and the head; the matched frozen arm updates the same head on fixed encoder features. Batch-normalization running moments remain fixed in both fine-tuning continuations. Calibration selects checkpoints by NLL and fits a scalar temperature. The study also evaluates source-level tree ensembles on the fully fine-tuned features, and probability consensus between the neural and tree outputs.

Corruption training samples one image view per source in an epoch or ensemble member, preserving the case label and pulse. Views include clean, dim, blurred, and combined images. Image-derived colour measurements are recomputed after corruption. Clean controls use the same training sources and fitting recipe without those photometric changes. Further training updates all encoder parameters. If calibration selects epoch zero, we retain the starting model.

\subsection{Likelihood model and posterior update}
For the decision-support experiments, $v$ is the temperature-calibrated mean of three ExtraTrees classifiers trained on the fully fine-tuned features with source-level corruption augmentation. Let $r(h\mid z)$ be a pulse-only reference trained in the same fold. The inquiry prior is
\begin{equation}
p_0(h)=0.95\,v(h\mid x,z)+0.05\,r(h\mid z).
\label{eq:prior}
\end{equation}
The reference component preserves some probability for alternatives when the multimodal classifier is concentrated. The question bank defines 32 atomic concepts with source-record identifiers and declared costs $c_q$. Smoothed counts from the training cases estimate the answer likelihood $L_{hqa}=P(A_q=a\mid h)$.

For a known acquired answer $a_t$ to concept $q_t$, the update is
\begin{equation}
p_{t+1}(h)=\frac{p_t(h)L_{h q_t a_t}}{\sum_j p_t(j)L_{j q_t a_t}}.
\label{eq:update}
\end{equation}
Unknown responses leave $p_t$ unchanged. The implementation maintains the two experts and their accumulated evidence weights; this is numerically equivalent to updating the mixture in Eq.~\ref{eq:prior}. The record links each answer and posterior update to the likelihood table and its source. The language model handles wording and answer extraction; Eqs.~\ref{eq:prior}--\ref{eq:update} determine the probabilities.

\subsection{Acquisition and the candidate check}
Expected information gain (EIG) connects the current visual-conditioned belief to question selection. Below the reporting threshold, acquisition maximizes expected entropy reduction per cost:
\begin{align}
I(q;p_t)&=H(p_t)-\sum_a P(a\mid q,p_t)H(p_t^{q,a}),\\
q_t&=\arg\max_{q\notin\mathcal{Q}_t} I(q;p_t)/c_q.
\label{eq:eig}
\end{align}
The posterior therefore controls both which question is asked and the value assigned to its possible answers. The total 32-concept cost is 33.2. We evaluate cost ceilings $0,2,4,6,8,10,12,16,20,24,28,33.2$. The budget curves use the affordable prefix of each policy's trajectory.

The static comparator greedily selects the next question by mean EIG per cost across training cases, updating their beliefs with recorded training answers after each selection. The resulting order is fixed during evaluation.

When candidate $h^*$ crosses its class-calibrated confidence threshold, the candidate check uses a different comparison. Let $\ell_{h,q}$ denote the three-outcome likelihood vector and $\bar\ell_{-h,q}$ the uniform average over the other 32 hypotheses. The check score is
\begin{equation}
V(q,h^*)=\frac{\operatorname{JS}(\ell_{h^*,q},\bar\ell_{-h^*,q})}{c_q}.
\label{eq:verify}
\end{equation}
JS denotes Jensen--Shannon divergence. The score selects questions that distinguish the leading candidate from the remaining hypotheses, even when those hypotheses have little posterior mass. Answers still update the posterior through Eq.~\ref{eq:update}. An ordinary report requires one additional known answer that leaves the same candidate above threshold. A change of candidate or loss of confidence resets confirmation. Each concept can be asked once.

\subsection{Detecting and resolving a discrepancy}
Let $h_0=\arg\max_h v(h\mid x,z)$ be the initial objective hypothesis. We monitor $s_t=1-p_t(h_0)$ after acquired observations. Calibration chooses an alert threshold controlling the maximum interview score to at most 10\% empirical first-alert rate among initially correct cases, separately across calibration image conditions and answer draws. At the alert time $t_*$, we save $p_{t_*}$ to measure the support gained during reassessment.

The review-pause baseline stops at the alert and refers the case for review. Active reassessment continues asking questions. A flagged case may produce a report only when a stable candidate remains above its class threshold through two subsequent known answers and satisfies
\begin{equation}
\frac{p_t(h)/(1-p_t(h))}{p_{t_*}(h)/(1-p_{t_*}(h))}\geq 10.
\label{eq:resolve}
\end{equation}
Only answers acquired after the alert count toward this requirement. Unknown answers leave the alert open. We retain the full alert history and test a stricter odds factor of 100 in the sensitivity analysis. Cases that exhaust their budget with an unresolved alert remain in review.

Before an alert, the policy follows the same one-answer check, so the alert calibration applies to both stopping and reassessment. The terminal outcomes are a report, a referral for review, or budget exhaustion. Safety screening takes priority over all three. Throughout reassessment, new answers update the existing posterior.

\subsection{Selective reporting and language boundary}
For each predicted fine class, calibration tests thresholds in $\{.90,.925,.95,.975,.99\}$. A threshold requires at least five distinct calibration cases and 95\% precision; otherwise that class's point report is disabled while it remains an inference candidate. Each source is counted once across image variants and answer draws. Reporting success on held-out cases requires at least 20 reports, at least 90\% precision, and a Wilson 95\% lower bound of at least 80\% in \emph{every} answer draw of \emph{every} fold. Family reports never count as fine resolution.

Local Qwen3:4b-instruct through Ollama phrases the selected canonical concept and extracts an answer from the latest reply. Phrasing can use dialogue history; extraction uses the current concept and current reply to avoid carrying an earlier answer into a new observation. A copied supporting quote is required. Ambiguity can trigger one clarification of the same concept, charged as another patient turn and cost without creating a second independent symptom observation. Canonical content and source identifiers are retained when generated wording is incomplete.

\section{Visual representation and classification}
\subsection{Controlled visual training intervention}
We first isolate visual representation quality with three training seeds and checkpoints at zero, one, five, and ten domain fine-tuning epochs. The pulse encoder, 600 downstream training cases, 200 development cases, logistic fine classifier recipe, question likelihoods, and answer draws are held fixed. Stage zero fits an attribute head on a frozen ImageNet backbone; later stages update all visual layers along the same training path. We evaluate all four checkpoints.

\begin{figure*}[t]
\centering\includegraphics[width=\textwidth]{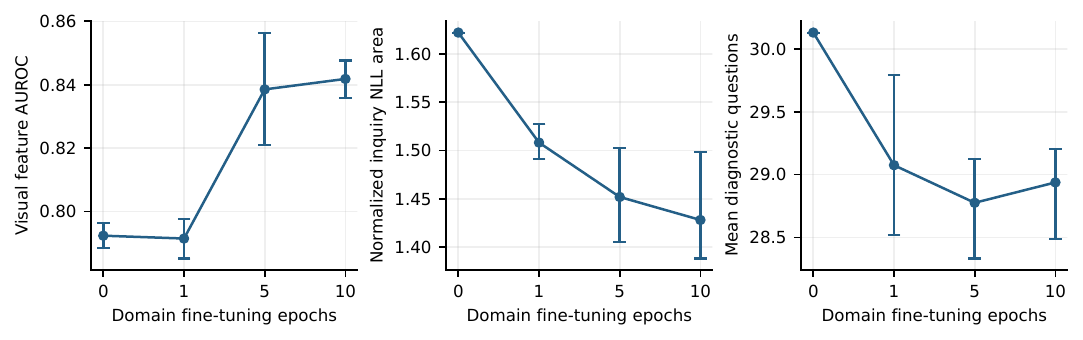}
\caption{Controlled visual intervention on a 200-case evaluation subset of the case bank. Results cover all four checkpoints and three training seeds. Points show means; ranges show variation across trained models. The fixed pulse encoder and acquisition engine isolate how visual training changes predictive evidence and subsequent inquiry. Its split differs from the cross-fit protocol, but the cases overlap.}
\label{fig:vision}
\end{figure*}

On the 200-case evaluation cohort, mean visual AUROC rises from 0.792 at stage zero to 0.842 at stage ten. Initial fine accuracy rises from 28.5\% to 36.8\%, and initial NLL falls from 2.338 to 2.083. The corresponding mean perplexity falls from 10.36 to 8.05. Across the complete acquisition frontier, normalized NLL area improves from 1.622 to 1.428; mean question count falls from 30.13 to 28.94. Stage ten changes the first selected question in 46.67\% of paired cases relative to stage zero. Figure~\ref{fig:vision} shows training-seed ranges alongside the mean.

Better image features change both the initial prediction and the questions that follow it. The best checkpoint depends on the metric: stage five has slightly higher initial fine accuracy and asks fewer questions than stage ten, while stage ten has lower NLL. The policy uses entropy and confidence online; NLL and perplexity are computed against the assigned labels for evaluation.

\subsection{Full multimodal fine-tuning}
The five-fold classifier comparison uses the same source partitions for the ExtraTrees baseline, frozen neural, and fully fine-tuned neural classifiers. The fully fine-tuned neural model reaches 50.25\% fine and 71.95\% family accuracy before inquiry; its matched frozen neural control reaches 50.02\% and 71.84\%. The neural training recipe improves held-out fine accuracy over the ExtraTrees baseline by 4.27 percentage points, with paired-case interval $[2.58,6.00]$. Unfreezing adds 0.23 points over the matched frozen neural head, with interval $[-0.77,1.19]$. Most of the gain comes from the neural training recipe; the interval for unfreezing includes zero.

Corruption-aware heads and further full-encoder continuation improve dim and blur discrimination. The neural model trained further on corrupted images reaches 66.77\% full-inquiry fine accuracy on dim images and 67.06\% under blur, compared with 61.92\% and 60.54\% for the clean-trained, fully fine-tuned neural baseline. These neural priors fail the reporting criterion in some folds despite their higher accuracy.

\section{Visual evidence, inquiry, and decision support}
\subsection{Controlled comparisons and endpoints}
The component comparison fixes the source ensemble, likelihoods, cases, answers, class thresholds and discrepancy threshold. Replaying the saved answers reproduces the posteriors. We rerun ordinary EIG for each prior intervention. Counterfactual budget curves continue all cases, whereas operational results follow actual report, review or exhaustion. Full-bank posteriors are order invariant for an unchanged prior, so acquisition quality is measured at intermediate costs. Paired intervals use 2,000 case-bootstrap samples after averaging answer draws within cases. 

\subsection{How predicted families help inquiry}
Let $P(h)$ be the initial 33-way mixture and $F_g=\sum_{h\in g}P(h)$ its family mass. Let $\pi(h)$ and $\pi_g$ denote smoothed training frequencies. Four priors isolate different information:
\begin{equation}
\begin{aligned}
 P_{\rm full}(h)&=P(h), & P_{\rm family}(h)&=F_g\pi(h\mid g),\\
 P_{\rm within}(h)&=\pi_gP(h\mid g), & P_{\rm none}(h)&=\pi(h).
\end{aligned}
\label{eq:family}
\end{equation}
Every arm retains all 33 candidates. The family-only arm preserves objective family evidence but removes case-specific distinctions within a family; the within-only arm makes the reverse intervention. This tests family information at inference separately from the auxiliary family training loss.

At cost 12 on clean cases, family-only inquiry reaches 47.41\% fine accuracy versus 34.40\% without objective evidence, a paired gain of $13.01$ ($[12.30,13.71]$) percentage points. Restoring case-specific within-family preferences raises accuracy to 65.78\%, a further $18.37$ ($[17.13,19.57]$) points. Removing predicted family masses from the full prior reduces accuracy by $15.68$ ($[14.92,16.43]$) points. Family information and within-family preferences therefore contribute separately to the inquiry trajectory (Table~\ref{tab:family}).

\begin{table}[t]\centering\caption{Fine accuracy (\%) at cost 12 under controlled family-prior interventions; ordinary EIG for every arm.}\label{tab:family}\begin{tabular}{lrrr}\toprule Prior&Clean&Dim&Blur\\\midrule
Full&65.78&64.49&65.53\\
Family only&47.41&46.85&47.66\\
Within only&50.09&49.70&49.88\\
No objective&34.40&34.40&34.40\\\bottomrule\end{tabular}\end{table}

\subsection{Differential narrowing across acquisition budgets}
Before inquiry, the full initial mixture has clean fine accuracy of 46.44\% and top-three inclusion of 77.26\%. At cost 12, these become 65.78\% and 90.75\%, with top-two inclusion of 83.91\%; matched static inquiry reaches 59.45\% and 87.34\%. Under dimming and blur, adaptive cost 12 top-three inclusion is 89.70\% and 91.09\%. After all 32 concepts, the assigned syndrome is among the top three candidates in 91--92\% of cases across the three primary image conditions. 

\begin{figure*}[t]
\centering\includegraphics[width=\textwidth]{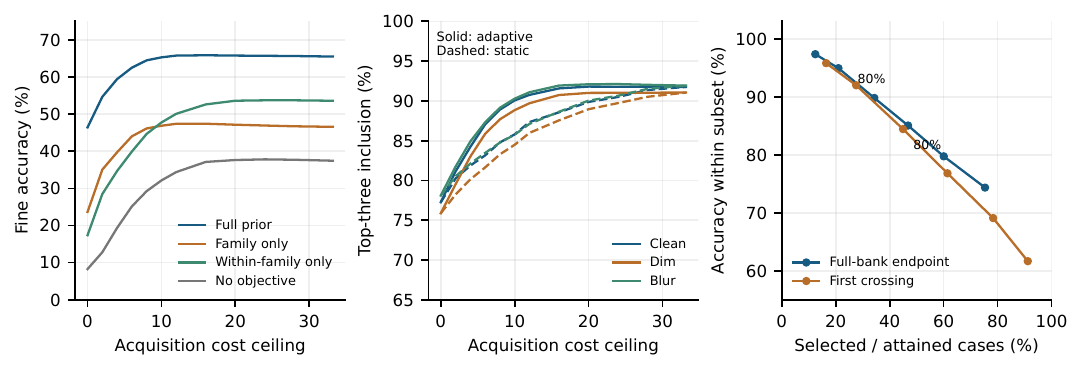}
\caption{Decision support from objective evidence and inquiry. Left: the controlled family-prior interventions under clean images. Middle: top-three inclusion for adaptive and static inquiry across image conditions. Right: clean-image confidence tradeoffs at the full-bank endpoint and the first threshold crossing. Points cover the full threshold grid.}
\label{fig:decision}
\end{figure*}

Adaptive-minus-static differences in normalized NLL--cost area are $-0.107$ ($[-0.113,-0.100]$) under dimming and $-0.105$ ($[-0.112,-0.099]$) under blur. Under shared reporting and review settings, mean operating cost differences are $-0.828$ ($[-0.928,-0.729]$) and $-0.813$ ($[-0.910,-0.716]$), respectively. 

\subsection{Confidence and early decisions}
The median full-bank leading probability is 66--68\% across clean, dim, and blur. A threshold of 80\% selects 32--34\% of cases with 89--90\% accuracy; a threshold of 90\% selects 19--21\% with approximately 95\% accuracy. Table~\ref{tab:confidence} separates this endpoint selection from stopping at the first confidence crossing. The latter reaches more cases earlier, with lower accuracy. At 80\%, clean first-crossing attainment is 44.94\%, accuracy is 84.43\%, and mean cost among attained cases is 3.73. Mean cost here is conditional on reaching the threshold.

\begin{table*}[t]\centering\caption{Clean-image confidence tradeoffs (\%). Full-bank subsets and first crossings are descriptive decision-support endpoints, without the operational reporting/review gates. First-crossing cost averages only attained cases.}\label{tab:confidence}\begin{tabular}{lrrrrr}\toprule &\multicolumn{2}{c}{After all 32 concepts}&\multicolumn{3}{c}{First threshold crossing}\\Threshold&Selected&Accuracy&Attained&Accuracy&Cost if attained\\\midrule
60\%&60.03&79.73&78.35&69.08&2.76\\
80\%&34.30&89.79&44.94&84.43&3.73\\
90\%&21.00&94.92&27.58&91.98&4.43\\
95\%&12.43&97.34&16.46&95.79&5.27\\\bottomrule\end{tabular}\end{table*}

The 60\% endpoint subset includes all higher-confidence cases. Within the narrower 60--70\% confidence band, accuracy is about 61--62\%. These curves describe confidence-based selection; operational reports also require the class-specific threshold and the candidate check.

\begin{figure*}[t]
\centering\includegraphics[width=\textwidth]{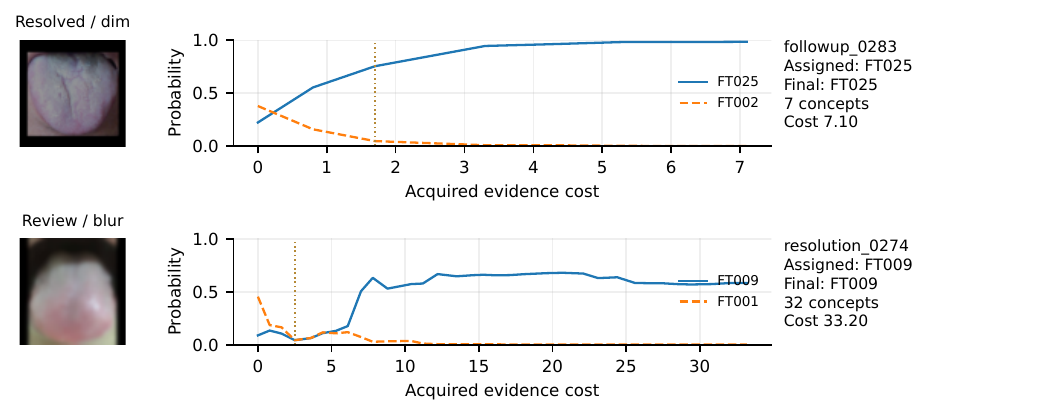}
\caption{Two hash-selected routes, with fixed answer draw 7101. Solid curves identify the assigned state only for retrospective display. Fresh supporting evidence closes the dim case's discrepancy; the blurred case yields an improved differential while retaining review.}
\label{fig:worked}
\end{figure*}

\subsection{What each component changes}
Table~\ref{tab:components} reports the controlled dim-image comparison. One-answer confirmation reduces both report coverage and erroneous-report frequency relative to immediate EIG reporting. The candidate check recovers $0.239$ ($[0.150,0.335]$) percentage points of coverage over local confirmation, with a cost change of $-0.080$ ($[-0.109,-0.053]$). Its normalized NLL-area change is $-0.0000006$ ($[-0.0000279,0.0000282]$); most of the acquisition-curve advantage over static order is already supplied by ordinary EIG. The candidate check therefore changes reporting more than acquisition quality.

\begin{table}[t]\centering\caption{Controlled components under dim images. Coverage and precision are percentages; cost follows actual stopping. Class and alert thresholds are shared across arms.}\label{tab:components}\begin{tabular}{lrrr}\toprule Component&Cov.&Prec.&Cost\\\midrule
Ordinary EIG&7.97&97.73&30.97\\
Local confirmation&7.38&98.02&31.24\\
Candidate check&7.61&97.93&31.16\\
Check + review pause&6.70&97.82&19.32\\
Check + reassessment&7.36&97.91&31.23\\
Static + reassessment&6.68&97.64&32.06\\\bottomrule\end{tabular}\end{table}

Immediate review stops acquisition at a discrepancy. Active reassessment instead obtains fresh evidence, increasing dim report coverage from 6.70\% to 7.36\%, a paired change of $0.658$ ($[0.504,0.831]$) percentage points, while mean cost rises from 19.32 to 31.23. Reports issued after resolving a dim-image alert have 98.83\% precision. The complete configuration passes the reporting criterion across five folds, three primary conditions, and ten answer draws; combined dim and blur passes four folds. The candidate check without reassessment also passes the primary precision and count checks. Reassessment adds a rule for resolving contradictory cases at the cost of longer interviews.

\subsection{Recorded routes and implementation}
Figure~\ref{fig:worked} shows the first hash-ranked case from a resolved dim-image case and an unresolved blurred-image case, selected without filtering for correctness. The first changes its leading hypothesis from FT002 to assigned FT025 and reports after seven concepts at cost 7.10. In the second, assigned FT009 becomes most probable but remains below its report threshold after 32 concepts. The record retains the photograph, posterior, asked concepts, posterior at the alert, and subsequent supporting answers.

We check the implementation with raw inputs and by replaying structured answers. A paired dialogue evaluation uses 12 hash-selected cases, cycling clean, dim and blur, with identical communication conditions and cost caps. Prompted Qwen recovers 98.07\% of assigned categories over 311 turns. Fine predictions are correct in 10/12 Qwen episodes versus 9/12 structured episodes; top-three inclusion is 11/12 versus 12/12. Mean acquisition costs, including clarification costs, are 25.96 and 29.68. Across the twelve episodes, the respective arms use 51 and 30 clarification turns. This separate dialogue experiment uses Qwen without fine-tuning.

\subsection{Sensitivity to the simulated answer model}
With the classifier and reporting rules fixed, withholding 20\% of answers yields 62.43\% full fine accuracy and 89.93\% top-three inclusion; reversing 20\% of answers yields 51.32\% and 80.49\%. Halving fine-specific symptom expression yields 59.12\% fine accuracy. A privileged likelihood analytically integrating the generator's patient variation raises baseline accuracy from 65.51\% to 66.60\%, a paired gain of $1.09$ ($[0.71,1.44]$) points. The reference integrates over patient variation without access to a test case's hidden state. Its small gain indicates that likelihood estimation accounts for only part of the remaining uncertainty. The supplement gives all perturbation results.

\section{Limitations and scope}
The benchmark uses real photographs and their visual annotations, combined with assigned syndromes, rendered pulses, and simulated symptom responses. The assigned syndromes are benchmark targets rather than verified diagnoses of the photographed people. The hierarchy and pulse mapping require independent assessment. Source clustering reduces image duplication, but person-level separation cannot be verified. FT019 is used only for training, and support for the other terms is unequal.

All findings come from retrospective development on one source bank. Fold-wise exclusion prevents direct fitting on evaluation cases, but repeated development may still favour this bank. The bootstrap intervals condition on fitted models and exclude uncertainty from model selection or new populations. The selected source ensemble reaches 65.51\% clean full-inquiry accuracy, compared with 68.15\% for the clean-trained, fully fine-tuned neural baseline. Neural and consensus alternatives improve discrimination but fail some reporting folds; combined dimming and blur also remains difficult.

The factorized likelihood can overcount correlated symptoms, and the question bank limits what the system can ask. Top-three inclusion measures the quality of the ranked candidates; a correct point report requires a stricter decision. Reassessment improves coverage over stopping at an alert, with a substantial increase in cost. Its check and odds thresholds are research choices, and the reported costs are assigned acquisition weights, not consultation time. Natural patient dialogue, clinical decision thresholds, and the sensitivity of safety screening need independent validation.

\section{Conclusion}
Tongue-image assessment can use visual findings to guide the search for symptoms that the image cannot reveal. In our constructed-case benchmark, combining image/pulse evidence with adaptive inquiry reaches 65.8\% fine-syndrome accuracy and 90.8\% top-three inclusion at cost 12, exceeding either evidence source alone and a same-prior static order. The 200-case visual intervention shows that image representation also changes which question is asked. Reporting controls address a distinct application need: deciding when new supporting answers justify a report after a discrepancy, with measured coverage and cost. Together, the results connect visual assessment to targeted evidence gathering and a recorded basis for reporting or review.

\clearpage\section*{LLM Usage Statement}
We used large language models (OpenAI's ChatGPT and Anthropic's Claude) in a limited, assistive way while preparing this work. The full manuscript, or major parts of it, was given to these models to get reviewer-style feedback. LLMs were also used to edit and refine text the authors wrote (language, clarity and structure), to edit and refine code the authors wrote, and to refine figures and visual material the authors made. LLMs did not generate research ideas, hypotheses, claims or conclusions. They also played no part in designing, running or analyzing experiments, in creating or annotating data, or in preparing related work and citations. The authors reviewed and verified every change an LLM suggested and take full responsibility for all submitted content.

Separately, Qwen3-4B-Instruct ran locally through Ollama inside the evaluated dialogue interface, where it phrased questions, extracted answers and produced simulated replies. It never assigned a diagnosis, chose a question or set a probability.

\clearpage
\onecolumn
\appendix
\setcounter{section}{0}
\renewcommand{\thesection}{S\arabic{section}}
\setcounter{table}{0}\renewcommand{\thetable}{S\arabic{table}}
\setcounter{figure}{0}\renewcommand{\thefigure}{S\arabic{figure}}
\section*{Supplementary Information}
\section{Evaluation scope and statistical unit}
The benchmark contains 2,600 constructed cases: 2,599 evaluated cases across five source-disjoint folds and one training-only case, FT019. All tables use the evaluated cohort unless stated otherwise. Each case has ten answer realizations, which are repeated measurements rather than independent cases. Paired 95\% intervals resample case identities 2,000 times after averaging answer draws within a case. Models are held fixed, so the intervals reflect case sampling only; they do not cover repeated development on the source bank or multiple comparisons.

The analyses are retrospective: their design was informed by development results and an exploratory confidence audit on the same photograph bank. The family-information intervention, generator perturbations and language audit were each specified before their respective outcomes were observed. Tables include configurations that do not meet the reporting criterion. These analyses do not constitute independent confirmation on an untouched source bank.

\section{Evidence for image-guided symptom inquiry}
The application question is how tongue and pulse findings can guide inquiry when symptoms needed for syndrome differentiation are missing. The matched-cost comparison tests whether adapting questions improves the shortlist beyond a fixed order with the same initial evidence. Family-prior interventions identify which parts of that evidence help inquiry. Component comparisons address a separate question: what can be reported when later answers weaken the initial assessment? They measure report precision, coverage and acquisition cost together. Generator perturbations test sensitivity to imperfect answers, and the paired language study examines delivery of the structured interview through dialogue.

The comparison below uses the same classifier, likelihood, cases, answers and reporting settings; only question ordering changes. It measures ranked prediction at a cost ceiling. Operational reporting additionally applies the class-specific confidence and review rules. The abstract rounds the clean-image results to one decimal place.
\begin{center}\small\begin{tabular}{lrrr}\toprule Images & Acquisition & Fine@12 & Top3@12\\\midrule
Clean & Adaptive & 65.78 & 90.75\\
Clean & Static & 59.45 & 87.34\\
Dim & Adaptive & 64.49 & 89.70\\
Dim & Static & 58.43 & 85.98\\
Blur & Adaptive & 65.53 & 91.09\\
Blur & Static & 59.35 & 87.11\\\bottomrule\end{tabular}\end{center}
\noindent\textit{Fine accuracy and top-three inclusion at acquisition cost ceiling 12, as percentages.}

\section{Controlled component outcomes}
The same objective classifier, trained likelihood, cases, answers and class/alert thresholds are used for every component intervention. Static acquisition changes only the question ordering: mean training-case EIG per cost selects each next question, training beliefs are updated with recorded answers, and the resulting order is fixed for evaluation. A fold passes when every answer draw yields at least 20 reports, at least 90\% precision and a Wilson 95\% lower bound of at least 80\%. Arms without a review gate can pass this count and still have no rule for contradictions. Ordinary EIG reports at its first class-specific confidence crossing. Local confirmation asks for one more supporting answer. The candidate check changes the question score at a crossing so that questions separate the candidate from all others. The review pause stops when posterior support for the initial image/pulse prediction crosses its calibrated alert threshold; reassessment continues and can close the alert once new answers support a stable candidate.
\begin{center}\small\begin{tabular}{lrrrrr}\toprule Component & Coverage & Precision & Cost & Review & Pass folds\\\midrule
Ordinary EIG & 9.11 & 97.72 & 30.62 & 0.00 & 4/5\\
Local confirmation & 8.42 & 97.94 & 30.95 & 0.00 & 4/5\\
Candidate check & 8.69 & 97.96 & 30.86 & 0.00 & 5/5\\
Check + review pause & 7.75 & 98.06 & 20.18 & 36.27 & 5/5\\
Check + reassessment & 8.34 & 97.97 & 30.96 & 35.68 & 5/5\\
Static + reassessment & 7.77 & 97.72 & 31.77 & 35.17 & 3/5\\\bottomrule\end{tabular}\end{center}
\noindent\textit{Clean operational outcomes. Coverage, precision and review fraction are percentages. Cost is the mean to actual report, review or exhaustion.}

\begin{center}\small\begin{tabular}{lrrrrr}\toprule Component & Coverage & Precision & Cost & Review & Pass folds\\\midrule
Ordinary EIG & 7.97 & 97.73 & 30.97 & 0.00 & 5/5\\
Local confirmation & 7.38 & 98.02 & 31.24 & 0.00 & 5/5\\
Candidate check & 7.61 & 97.93 & 31.16 & 0.00 & 5/5\\
Check + review pause & 6.70 & 97.82 & 19.32 & 39.86 & 4/5\\
Check + reassessment & 7.36 & 97.91 & 31.23 & 39.20 & 5/5\\
Static + reassessment & 6.68 & 97.64 & 32.06 & 39.00 & 4/5\\\bottomrule\end{tabular}\end{center}
\noindent\textit{Dim operational outcomes. Coverage, precision and review fraction are percentages. Cost is the mean to actual report, review or exhaustion.}

\begin{center}\small\begin{tabular}{lrrrrr}\toprule Component & Coverage & Precision & Cost & Review & Pass folds\\\midrule
Ordinary EIG & 9.02 & 98.08 & 30.63 & 0.00 & 5/5\\
Local confirmation & 8.39 & 98.49 & 30.94 & 0.00 & 5/5\\
Candidate check & 8.62 & 98.26 & 30.86 & 0.00 & 5/5\\
Check + review pause & 7.68 & 98.25 & 20.30 & 35.82 & 5/5\\
Check + reassessment & 8.27 & 98.28 & 30.95 & 35.23 & 5/5\\
Static + reassessment & 7.85 & 98.14 & 31.77 & 35.05 & 4/5\\\bottomrule\end{tabular}\end{center}
\noindent\textit{Blur operational outcomes. Coverage, precision and review fraction are percentages. Cost is the mean to actual report, review or exhaustion.}

\begin{center}\small\begin{tabular}{lrrrrr}\toprule Component & Coverage & Precision & Cost & Review & Pass folds\\\midrule
Ordinary EIG & 7.57 & 97.92 & 31.07 & 0.00 & 5/5\\
Local confirmation & 7.06 & 98.15 & 31.33 & 0.00 & 4/5\\
Candidate check & 7.20 & 98.02 & 31.27 & 0.00 & 4/5\\
Check + review pause & 6.28 & 97.79 & 19.16 & 40.80 & 4/5\\
Check + reassessment & 6.93 & 97.95 & 31.35 & 40.15 & 4/5\\
Static + reassessment & 6.50 & 98.28 & 32.06 & 40.02 & 4/5\\\bottomrule\end{tabular}\end{center}
\noindent\textit{Combined dim + blur operational outcomes. Coverage, precision and review fraction are percentages. Cost is the mean to actual report, review or exhaustion.}

\section{Correction, harm, and component contrasts}
Correction means an initially wrong prediction becomes correct; harm means an initially correct prediction becomes wrong. These rates use the corresponding initial-correctness stratum. Operating predictions at a review stop are recorded for analysis and are not counted as issued diagnoses.
\begin{center}\small\begin{tabular}{lrr}\toprule Condition / component & Correction & Harm\\\midrule
Clean / Ordinary EIG & 45.91 & 11.78\\
Clean / Check + review pause & 37.44 & 14.27\\
Clean / Check + reassessment & 45.91 & 11.81\\
Clean / Static + reassessment & 45.90 & 11.82\\
Dim / Ordinary EIG & 45.80 & 12.30\\
Dim / Check + review pause & 36.87 & 15.05\\
Dim / Check + reassessment & 45.80 & 12.31\\
Dim / Static + reassessment & 45.80 & 12.30\\
Blur / Ordinary EIG & 45.25 & 11.44\\
Blur / Check + review pause & 36.81 & 13.76\\
Blur / Check + reassessment & 45.25 & 11.44\\
Blur / Static + reassessment & 45.24 & 11.44\\\bottomrule\end{tabular}\end{center}
\noindent\textit{Correction and harm at actual operating states, as percentages within each initial-correctness stratum.}

\begin{center}\small\begin{tabular}{lrrr}\toprule Contrast & Coverage & Wrong reports & Cost\\\midrule
Clean / Confirm $-$ EIG & -0.693 [-0.816, -0.573] & -0.035 [-0.069, -0.008] & 0.329 [0.287, 0.374]\\
Clean / Check $-$ local & 0.269 [0.177, 0.362] & 0.004 [-0.019, 0.027] & -0.094 [-0.121, -0.066]\\
Clean / Reassess $-$ pause & 0.593 [0.458, 0.739] & 0.019 [0.004, 0.038] & 10.778 [10.338, 11.261]\\
Dim / Confirm $-$ EIG & -0.593 [-0.700, -0.489] & -0.035 [-0.062, -0.012] & 0.270 [0.234, 0.308]\\
Dim / Check $-$ local & 0.239 [0.150, 0.335] & 0.012 [-0.012, 0.038] & -0.080 [-0.109, -0.053]\\
Dim / Reassess $-$ pause & 0.658 [0.504, 0.831] & 0.008 [0.000, 0.019] & 11.907 [11.472, 12.379]\\
Blur / Confirm $-$ EIG & -0.635 [-0.743, -0.519] & -0.046 [-0.077, -0.019] & 0.306 [0.266, 0.346]\\
Blur / Check $-$ local & 0.235 [0.146, 0.323] & 0.023 [0.004, 0.046] & -0.079 [-0.105, -0.054]\\
Blur / Reassess $-$ pause & 0.593 [0.458, 0.739] & 0.008 [0.000, 0.019] & 10.655 [10.222, 11.119]\\\bottomrule\end{tabular}\end{center}
\noindent\textit{Paired differences with case-bootstrap intervals. Coverage and wrong-report fractions use percentage points of all cases, not conditional precision.}

\section{Family-information intervention}
The full prior mixes the objective posterior with the pulse-only reference using weights 0.95 and 0.05. Family-only preserves its family masses and substitutes training-frequency conditionals. Within-only preserves its within-family conditionals and substitutes training family masses. No-objective uses training fine frequencies. Every arm uses ordinary EIG, the same trained likelihood, all 33 candidates and the same answers. This is an inference-information intervention, not an ablation of auxiliary family-loss training.
\begin{center}\small\begin{tabular}{lrrrrr}\toprule Condition / prior & Fine@12 & Top3@12 & Full fine & NLL area & Q1 changed\\\midrule
Clean / full & 65.78 & 90.75 & 65.51 & 1.065 & 0.00\\
Clean / family only & 47.41 & 84.51 & 46.58 & 1.501 & 39.48\\
Clean / within-family only & 50.09 & 74.09 & 53.61 & 1.757 & 82.84\\
Clean / no image/pulse evidence & 34.40 & 64.24 & 37.46 & 2.229 & 94.88\\
Dim / full & 64.49 & 89.70 & 64.47 & 1.123 & 0.00\\
Dim / family only & 46.85 & 83.65 & 46.16 & 1.534 & 40.32\\
Dim / within-family only & 49.70 & 73.68 & 53.13 & 1.785 & 84.42\\
Dim / no image/pulse evidence & 34.40 & 64.24 & 37.46 & 2.229 & 93.96\\
Blur / full & 65.53 & 91.08 & 65.24 & 1.069 & 0.00\\
Blur / family only & 47.66 & 84.97 & 46.73 & 1.488 & 41.55\\
Blur / within-family only & 49.88 & 73.92 & 53.03 & 1.774 & 84.38\\
Blur / no image/pulse evidence & 34.40 & 64.24 & 37.46 & 2.229 & 95.61\\
Combined dim + blur / full & 62.57 & 88.70 & 62.53 & 1.189 & 0.00\\
Combined dim + blur / family only & 46.20 & 82.92 & 45.77 & 1.556 & 43.29\\
Combined dim + blur / within-family only & 48.56 & 72.97 & 51.72 & 1.834 & 85.88\\
Combined dim + blur / no image/pulse evidence & 34.40 & 64.24 & 37.46 & 2.229 & 94.34\\\bottomrule\end{tabular}\end{center}
\noindent\textit{Accuracy and first-question change are percentages. NLL area is normalized by the maximum cost 33.2. These arms use ordinary EIG; the matched-cost table uses active reassessment, whose candidate-check ordering can produce slightly different intermediate-budget results.}

\section{Confidence: endpoint versus first crossing}
We evaluate confidence thresholds of 50, 60, 70, 80, 90 and 95\%. Endpoint subsets use the posterior after all 32 concepts. First crossing includes zero-cost crossings at the initial belief; if no crossing occurs, cost is recorded at exhaustion. These descriptive endpoints skip the class-disable, confirmation and review gates, so they are not operational coverage. Every threshold appears in the following tables.
\begin{center}\small\begin{tabular}{lrrrrrr}\toprule Threshold & End subset & End acc. & First attained & First acc. & Cost attained & Cost all\\\midrule
50 & 75.33 & 74.34 & 91.19 & 61.66 & 2.09 & 4.84\\
60 & 60.03 & 79.73 & 78.35 & 69.08 & 2.76 & 9.35\\
70 & 46.81 & 85.03 & 61.41 & 76.80 & 3.29 & 14.83\\
80 & 34.30 & 89.79 & 44.94 & 84.43 & 3.73 & 19.95\\
90 & 21.00 & 94.92 & 27.58 & 91.98 & 4.43 & 25.27\\
95 & 12.43 & 97.34 & 16.46 & 95.79 & 5.27 & 28.60\\\bottomrule\end{tabular}\end{center}
\noindent\textit{Clean: subset fractions and accuracy are percentages. First-crossing conditional cost excludes nonattainment; the last column includes it at exhaustion.}

\begin{center}\small\begin{tabular}{lrrrrrr}\toprule Threshold & End subset & End acc. & First attained & First acc. & Cost attained & Cost all\\\midrule
50 & 72.99 & 74.33 & 89.12 & 61.80 & 2.27 & 5.64\\
60 & 58.11 & 79.78 & 75.88 & 68.80 & 2.91 & 10.22\\
70 & 44.86 & 84.92 & 59.50 & 76.68 & 3.45 & 15.50\\
80 & 32.31 & 89.64 & 42.77 & 84.39 & 3.99 & 20.70\\
90 & 19.25 & 94.60 & 25.23 & 91.56 & 4.69 & 26.01\\
95 & 11.09 & 97.33 & 14.81 & 95.69 & 5.29 & 29.07\\\bottomrule\end{tabular}\end{center}
\noindent\textit{Dim: subset fractions and accuracy are percentages. First-crossing conditional cost excludes nonattainment; the last column includes it at exhaustion.}

\begin{center}\small\begin{tabular}{lrrrrrr}\toprule Threshold & End subset & End acc. & First attained & First acc. & Cost attained & Cost all\\\midrule
50 & 75.40 & 74.23 & 90.80 & 62.35 & 2.07 & 4.93\\
60 & 60.52 & 79.05 & 78.21 & 69.04 & 2.70 & 9.34\\
70 & 47.06 & 83.99 & 62.09 & 76.49 & 3.29 & 14.63\\
80 & 33.97 & 89.17 & 45.03 & 83.44 & 3.77 & 19.95\\
90 & 20.33 & 94.55 & 26.53 & 91.88 & 4.39 & 25.56\\
95 & 11.97 & 97.46 & 15.76 & 95.97 & 5.24 & 28.80\\\bottomrule\end{tabular}\end{center}
\noindent\textit{Blur: subset fractions and accuracy are percentages. First-crossing conditional cost excludes nonattainment; the last column includes it at exhaustion.}

\section{Generator sensitivity and privileged reference}
The photograph, pulse, assigned syndrome, objective model and reporting parameters remain fixed. Fine-specific expression is attenuated by moving each private symptom axis toward its family mean with factors 0.75 and 0.50. Missingness independently withholds 10\% or 20\% of concepts; unknown answers consume cost without updating or confirming. Answer reversal swaps absent and marked with probability 0.10 or 0.20, leaving the middle category unchanged. Baseline answer reconstruction exactly reproduces the saved answer matrices.

The generator reference analytically integrates the known clipped Gaussian symptom variation and linear answer-emission probabilities to obtain $P(a\mid h)$. It never receives a test case's hidden symptom vector or assigned syndrome at inference. It uses knowledge only the simulator has, so it is a reference point rather than a clinical likelihood or an accuracy ceiling. Its reporting thresholds are fixed to those of the learned-likelihood system, without recalibration, and the reporting numbers below are diagnostic only.
\begin{center}\small\begin{tabular}{lrrrrr}\toprule Condition & Full fine & Full top3 & Report cov. & Report prec. & Cost\\\midrule
Baseline & 65.51 & 91.76 & 8.34 & 97.97 & 30.96\\
Symptom expression $\times$0.75 & 62.36 & 90.02 & 7.40 & 97.14 & 31.21\\
Symptom expression $\times$0.50 & 59.12 & 88.49 & 6.61 & 96.04 & 31.43\\
10\% answers withheld & 64.17 & 90.84 & 7.32 & 97.95 & 31.24\\
20\% answers withheld & 62.43 & 89.93 & 6.23 & 97.71 & 31.52\\
10\% answers reversed & 58.65 & 86.76 & 6.86 & 97.03 & 31.33\\
20\% answers reversed & 51.32 & 80.49 & 5.76 & 95.06 & 31.62\\
Generator likelihood & 66.60 & 92.32 & 7.47 & 98.09 & 31.10\\
Generator likelihood, no image/pulse prior & 39.45 & 72.12 & 0.04 & 100.00 & 33.19\\\bottomrule\end{tabular}\end{center}
\noindent\textit{Answer perturbations and generator references under clean images. Reports require the reassessment gate; generator-reference thresholds are not recalibrated.}

\section{Paired language-interface evaluation}
Twelve case IDs were selected by a fixed hash, independently of correctness and reporting route, then assigned clean/dim/blur cyclically. One fixed answer draw was used. Both arms use the same active-reassessment pipeline, a 33.2 cost cap and an explicitly negative seven-item safety screen. The structured arm receives exact answer/clarification categories; Qwen receives only canonical concepts and natural replies. Case/question hashes inject 10\% refusal and 10\% initially unspecified replies in both arms. Each clarification consumes another turn and concept cost.
Qwen3:4b-instruct is prompted, not fine-tuned. Isolated private-patient calls realize assigned categories as natural Chinese; interviewer calls never receive categories, syndrome labels or objective probabilities. Question wording retains the canonical concept; extraction uses only the current reply. Category recovery measures the combined realization/extraction process. Using the same model for both roles, and only twelve episodes, limits what this audit can show. No prompt or parameter was changed during the audit.
\begin{center}\small\begin{tabular}{lrrrrrrr}\toprule Arm & Correct & Top3 & Turns & Clarify & Mean cost & Reports & Review\\\midrule
Structured & 9/12 & 12/12 & 351 & 30 & 29.68 & 2 & 4\\
Qwen & 10/12 & 11/12 & 311 & 51 & 25.96 & 3 & 4\\\bottomrule\end{tabular}\end{center}
\noindent\textit{Paired episode outcomes; report and pending-review columns are counts.}

Assigned-category recovery is 98.07\% over 311 parsed turns. Unsupported positive quote count is 0.
\begin{center}\small\begin{tabular}{lrrrrr}\toprule Case & Condition & Structured & Qwen & Cost diff. & Posterior L1\\\midrule
resolution\_0409 & Clean & 1 & 1 & 0.00 & 0.000\\
case\_0545 & Dim (0.6) & 0 & 0 & -0.10 & 0.212\\
case\_0584 & Blur ($\sigma{=}4$) & 1 & 1 & 1.20 & 0.051\\
resolution\_0343 & Clean & 1 & 1 & -24.00 & 0.021\\
extension\_0367 & Dim (0.6) & 1 & 1 & 0.60 & 0.104\\
extension\_0548 & Blur ($\sigma{=}4$) & 0 & 1 & 0.20 & 0.727\\
case\_0813 & Clean & 1 & 1 & -24.10 & 0.047\\
case\_0957 & Dim (0.6) & 1 & 1 & 0.40 & 0.207\\
resolution\_0387 & Blur ($\sigma{=}4$) & 1 & 1 & 0.10 & 0.070\\
followup\_0068 & Clean & 0 & 0 & 0.00 & 0.235\\
followup\_0024 & Dim (0.6) & 1 & 1 & 0.60 & 0.252\\
case\_0839 & Blur ($\sigma{=}4$) & 1 & 1 & 0.40 & 0.045\\\bottomrule\end{tabular}\end{center}
\noindent\textit{All selected episodes, including failures. Cost difference is Qwen minus structured; posterior difference is L1.}

\section{Classifier and robustness comparisons}
Separate pretraining uses 1,000 tongue training photographs plus 200 development photographs and 800 pulse training examples plus 200 development examples. Full multimodal fine-tuning updates every encoder stage and the head. Clean objective fine/family accuracies are 45.98/65.76\% for the ExtraTrees baseline, 50.02/71.84\% for the matched frozen neural model, and 50.25/71.95\% for full fine-tuning. The latter adds 0.23 fine-accuracy points over frozen, with paired interval [-0.77, 1.19]. Thus additional encoder updating was tested but its incremental gain was not established.
The decision-support source ensemble averages three ExtraTrees models (100 trees each, minimum leaf size 3, feature fraction 0.7) on fully fine-tuned features. Each member uses one independently sampled image view per training source; a scalar temperature calibrates the mean probabilities. This source ensemble reaches 65.51\% clean full-inquiry accuracy versus 68.15\% for the clean-trained, fully fine-tuned neural baseline. Neural reassessment reaches 66.77\% under dimming and 67.06\% under blur but passes 4/5 and 3/5 reporting folds, respectively. The comparison reflects a classification--reporting tradeoff. Supporting evidence reports corroboration, evidence-balance, normalization and consensus configurations, including those that do not meet the reporting criterion. All comparisons use the same source bank and are internal evaluations.
\section{Reproduction and case records}
Question concepts link through source-record identifiers to syndrome manifestations extracted from a supplied revision of \emph{TCM Clinical Diagnostic Terminology, Part 2: Syndromes}. Pulse descriptions come from transcribed reference pages. The source registry retains input hashes; answer probabilities are estimated from the constructed training cases.

Each case record links its assigned syndrome, source photograph, image-derived mask, rendered pulse array, private symptom state and fold assignment. Requested findings, source-image annotations and encoder predictions remain separate fields. The casebook shows syndrome labels, photographs, pulse heatmaps and the corresponding 18-channel waveforms together. Reproduction uses the saved folds, independently pretrained encoders, multimodal checkpoints, calibrated source ensemble, answer likelihoods and recorded trajectories. An evidence manifest records the input hashes of the component, prior, generator and language studies.

\end{document}